\documentclass[10pt,twocolumn,letterpaper]{article}

\usepackage{cvpr}
\usepackage{times}
\usepackage{epsfig}
\usepackage{graphicx}
\usepackage{amsmath}
\usepackage{amssymb}

\usepackage[                 breaklinks=true,                            bookmarks=false]{hyperref}

\cvprfinalcopy 

\def\cvprPaperID{3415} 

\begin{document}

\title{A new stereo formulation not using pixel and disparity models}

\author{Kiyoshi Oguri\ \ \ \ \ \ Yuichiro Shibata\\
Nagasaki University\\
Nagasaki, Japan\\
{\tt\small oguri,shibata@cis.nagasaki-u.ac.jp}
}

\maketitle

\begin{abstract}
  We introduce a new stereo formulation which does not use pixel and disparity models.
  Many problems in vision are treated as assigning each pixel a label.
  Disparities are labels for stereo.
  Such pixel-labeling problems are naturally represented in terms of energy minimization,
  where the energy function has two terms: one term penalizes solutions that inconsistent with the observed data,
  the other term enforces spatial smoothness.
  Graph cuts are one of the efficient methods for solving energy minimization.
  However, exact minimization of multi labeling problems can be performed
  by graph cuts only for the case with convex smoothness terms.
  In pixel-disparity formulation, convex smoothness terms do not generate well reconstructed 3D results.
  Thus, truncated linear or quadratic smoothness terms, etc. are used,
  where approximate energy minimization is necessary.
  In this paper, we introduce a new site-labeling formulation,
  where the sites are not pixels but lines in 3D space,
  labels are not disparities but depth numbers.
  For this formulation, visibility reasoning is naturally included in the energy function.
  In addition, this formulation allows us to use a small smoothness term,
  which does not affect the 3D results much.
  This makes the optimization step very simple, so we could develop
  an approximation method for graph cut itself (not for energy minimization) and
  a high performance GPU graph cut program.
  For Tsukuba stereo pair in Middlebury data set, we got the result in 5ms using GTX1080GPU,
  19ms using GTX660GPU.
\end{abstract}

\section{Introduction}
If we define stereo as a pixel disparity labeling problem,
this labeling becomes very difficult, since
disparities tend to be piecewise smooth.
They vary smoothly on the surface of an object,
but change dramatically at object boundaries.
Therefore discontinuity preserving property is necessary.
The goal is to find a labeling $f$ where $f$ is both piecewise smooth
and consistent with the observed data.
This problem can be naturally formulated in terms of energy minimization.
We seek the labeling $f$ that minimizes the energy
\[E(f) = E_{data}(f) + E_{smooth}(f),\]
where $E_{data}$ shows an energy corresponding to how the labeling $f$
consists with the observed data, while $E_{smooth}$ evaluates the smoothness of $f$.
The choice of $E_{smooth}$ is a critical issue and
many different functions have been proposed~\cite{Ajanthan,Boykov1,Scharstein}.
Also, this choice requires different optimization methods~\cite{Boykov3,Boykov2,Hirschmuller,Szeliski}.
Many basic methods for stereo use scalar(1D) disparity labels.
Such methods often implicitly assume front-parallel planes.
For example, standard piecewise smooth(\eg truncated linear or quadratic) pairwise regularization potentials
assign higher cost to surface with larger tilt.
To model surfaces more accurately Birchfield and Tomasi~\cite{Birchfield} introduced 3D-labels
corresponding to arbitrary 3D planes, but this approach is limited to piecewise planar scenes.
Woodford \etal~\cite{Woodford} retain the scalar disparity labels
while using triple-cliques to penalize 2nd derivatives of the reconstructed surface.
This encourages near planar smooth disparity maps. The optimization
problem is however made substantially more difficult due to the introduction of non-submodular
triple interactions.
Furthermore, volumetric graph cuts~\cite{Vogiatzis} and 3D-label energy model~\cite{Olsson} have been proposed.

In this paper we discard pixel-disparity model
which does not treat left and right image symmetrically.
We treat left and right image symmetrically
and introduce a novel concept of gaze lines, which are auxiliary lines defined on
the cross points of rays corresponding to left and right image pixels.
In our formulations, sites are gaze lines and labels are depth numbers.
For this simple formulation, visibility reasoning is naturally included in the energy function,
and a small smoothness term does not damage the 3D results much.
Therefore, we use convex function for smoothness and visibility terms.
This encourages exact optimization by graph cuts~\cite{Ishikawa2,Kolmogorov2}.
Our graph structure is some what similar to those of ~\cite{Ishikawa1}.

\begin{figure}[t]
\begin{center}
\fbox{
  \includegraphics[width=0.9\linewidth,clip,trim=50 50 100 40]{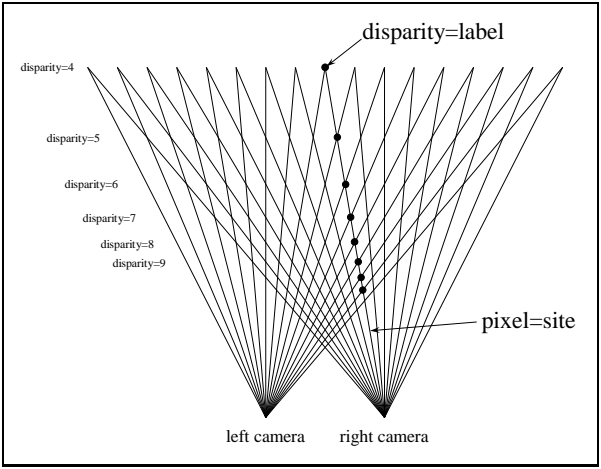}
}
\end{center}
\caption{Conventional pixel disparity model.}
\label{fig:1}
\end{figure}

\section{Energy model}
We define a first-order priors site-labeling problem as follows.
This assigns every site $v$ a label, which we write as $X_v$.
The collection of all site-label assignments is denoted by $X$.
The set of all site $v$ is denoted by $V$, so $v\in V$.
The set of all label $l$ is denote by $L$, so $X_v\in L$.
The number of sites is $n$, and the number of labels is $m$.
First order priors are defined on two-sites neighborhoods $(u,v)\in E\subset V\times V$.
Here $E$ denotes the set of all neighboring sites pairs.
The energy function $E(X)$ is
\[E(X)= \sum_{v\in V}g_v(X_v) + \sum_{(u,v)\in E}h_{uv}(X_u,X_v),\]
where $g_v(X_v)$ is called data term which penalizes solutions
that are inconsistent with the observed data,
and $h_{uv}(X_u,X_v)$ is pairwise potential (\ie interaction cost),
which includes smoothness energy and visibility reasoning energy~\cite{Woodford,Kolmogorov1}.
We use the standard 4-connected neighborhood system.
If the labels have a linear ordering and the interaction cost is an arbitrary convex function,
the problem can be solved exactly with graph cuts~\cite{Ishikawa2}.
These conditions are represented as
\[L=\{l_0,\dots,l_{m-1}\}\]
and
\[h_{uv}(l_i,l_j)=\tilde{h}_{uv}(i-j)\]
that is, labels form a line, and if
\begin{equation}
\tilde{h}_{uv}(i+1)-2\tilde{h}_{uv}(i)+\tilde{h}_{uv}(i-1)\ge 0\label{eq:convex},
\end{equation}
then interaction cost becomes convex function.

\section{New site and label}
\begin{figure*}
\begin{center}
\fbox{
  \includegraphics[width=0.7\linewidth,clip,trim=100 50 50 60]{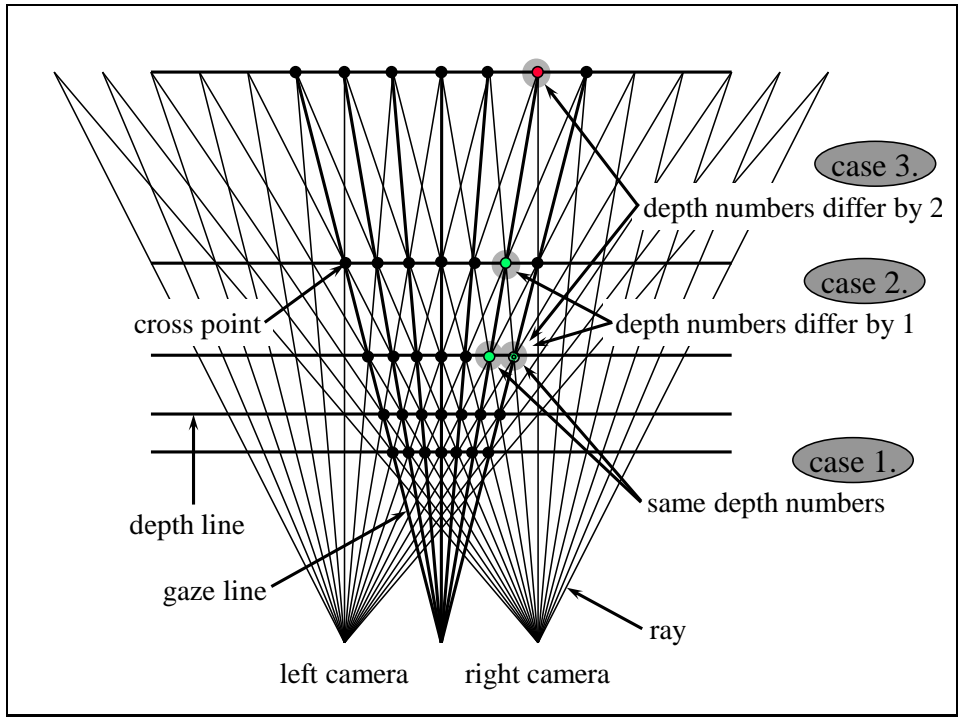}
}
\end{center}
\caption{Our new `gaze line' `depth number' model.
Gaze lines start the center of both eyes, \ie right and left cameras.
We think that I am in the center of both eyes. This figure also illustrates `cross points'.
The `real cross points' exist on the object surface.
}
\label{fig:2}
\end{figure*}

Figure~\ref{fig:1} shows the conventional pixel-disparity labeling,
where pixels are sites and disparities are labels.
Figure~\ref{fig:2} shows `cross points' of rays correspond to left and right image pixels
and `gaze lines' which connect some of the cross points.
We denote a coordinate of a left pixel as $(x_l,y_l)$, a right pixel as $(x_r,y_r)$, here
$x_l, x_r\in\{0,1,2,\dots,w-1\}$,
$y_l, y_r\in\{0,1,2,\dots,h-1\}$. The $(w,h)$ is image 2D size.
A cross point is represented as $(x_l,x_r,y)\in\{((x_l,y_l),(x_r,y_r))|y=y_l=y_r\}$.
A gaze lines is represented as
$(g,y)\in\{(x_l,x_r,y)|x_r+x_l=w-1+2g\}$, where $g$ is an integer and $-w/2 < g < w/2$.
A depth line is represented as
$(d,y)\in\{(x_l,x_r,y)|x_r-x_l=-(w-1)+2d\}$, where $d$ is an integer and $ 0 \le d < w/2$.
We call $d$ a depth number.
A cross point $(x_l,x_r,y)$ is also represented as $(g,d,y)$ by using $g$ and $d$.
These gaze lines form sites and depth numbers become labels.

\begin{figure*}
\begin{center}
\fbox{
  \includegraphics[width=0.7\linewidth,clip,trim=100 50 50 50]{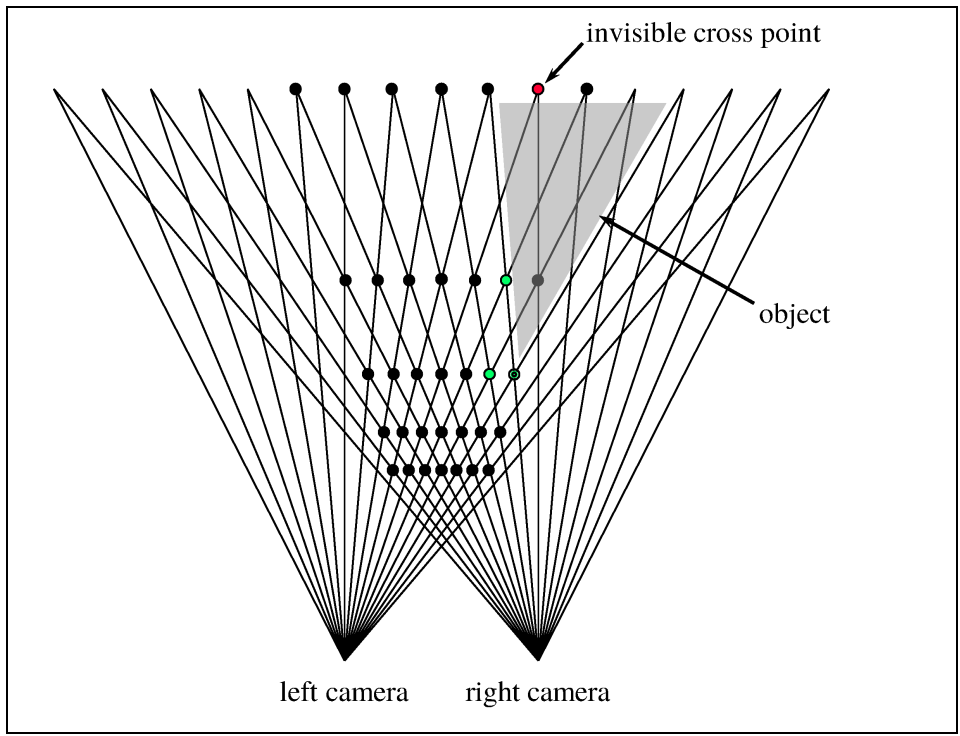}
}
\end{center}
\caption{The posterior `cross point' is behind some front object.}
\label{fig:3}
\end{figure*}

\begin{figure*}
\begin{center}
\fbox{
  \includegraphics[width=0.7\linewidth,clip,trim=20 40 40 40]{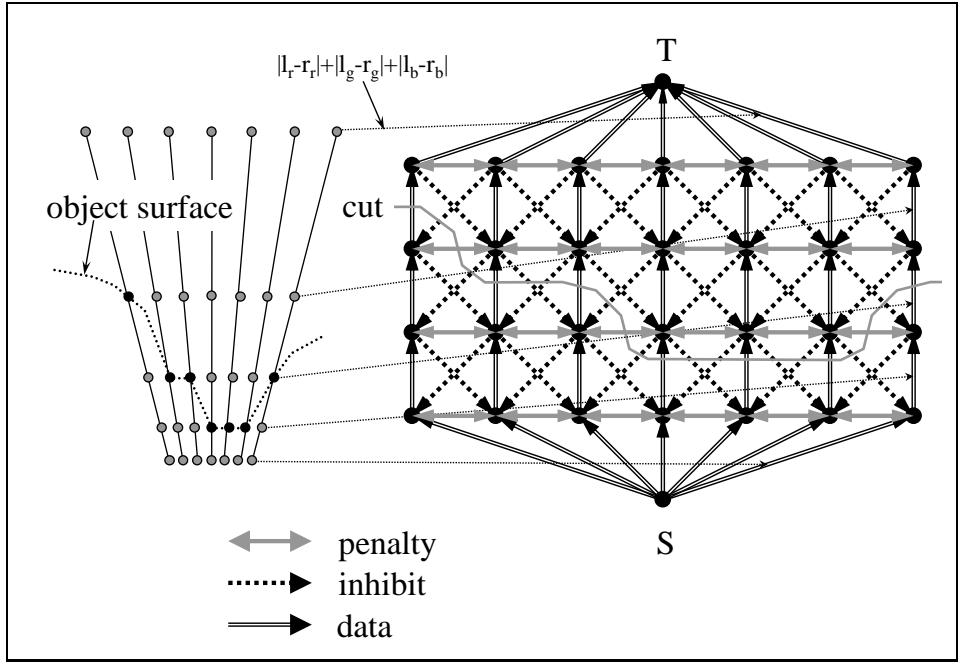}
}
\end{center}
\caption{The graph structure  representing the energy function.
Left portion indicates `gaze lines' and `depth numbers',
\ie `cross points', right portion indicates a corresponding graph structure.
`Cross points' correspond to the edges of the graph. 
This figure shows only one epipolar slice.
Actual graph is three dimensional.}
\label{fig:4}
\end{figure*}

Next we consider three cases (Figure~\ref{fig:2}).
We can assign\\
~1. the same depth numbers,\\
~2. two depth numbers differ by one,\\
~3. two depth numbers differ by more than one\\
between two neighboring sites.
Assigning the same depth numbers means a front parallel plane,
as same as in pixel-disparity model.
Assigning depth numbers differ by one means that the two cross points are on the same ray of the right or left camera.
Assigning depth numbers differ by more than one means that
the cross point in the back is invisible,
since it is behind some front object (Figure~\ref{fig:3}).
That is, assigning labels differ by more than one to neighboring sites
is inhibited in our site-labeling.
Therefore, we can write
\[\tilde{h}_{uv}(i-j)=\tilde{h}_1\times|i-j|+\tilde{h}_2\times(|i-j|-1)\times T(|i-j|>1).\]
Here, $i$ and $j$ are depth numbers,
$T(b)$ becomes 1 if $b$ is true, otherwise 0,
$\tilde{h}_1$ and $\tilde{h}_2$ are constant values,
and $\tilde{h}_2$ is infinity (actually enough big integer).
We call $\tilde{h}_1$ `penalty' for smoothness, $\tilde{h}_2$ `inhibit' for visibility.
This $\tilde{h}_{uv}(i-j)$ satisfies Equation~\ref{eq:convex}.
Therefore two-sites neighboring terms become a convex function.
So, we can minimize the energy by simple graph cuts.
This energy function is easily mapped to the graph structure
illustrated in Figure~\ref{fig:4}.

\begin{figure*}
\begin{center}
\fbox{
  \includegraphics[width=0.7\linewidth,clip,trim=30 50 50 60]{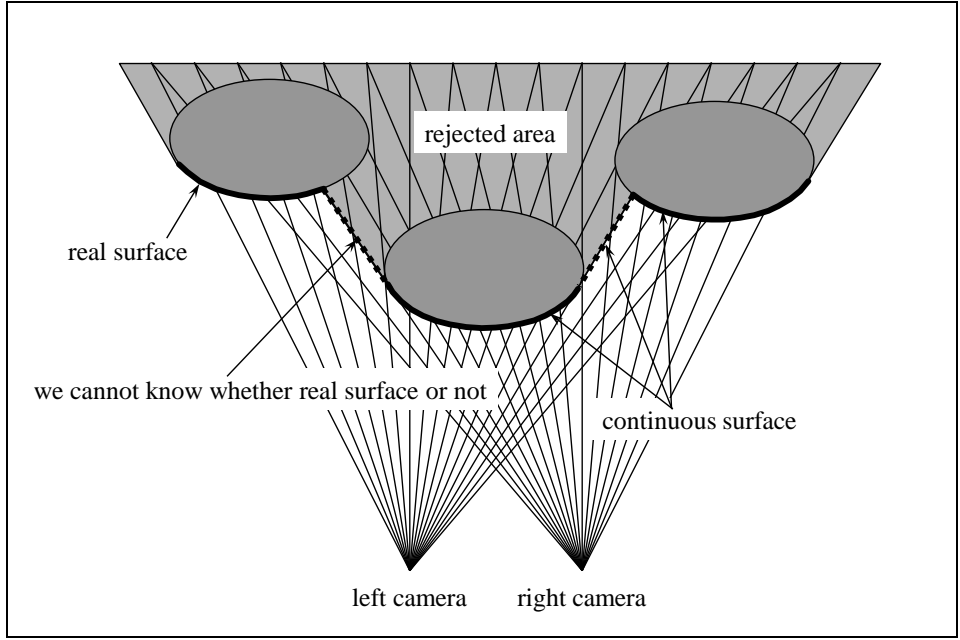}
}
\end{center}
\caption{We seek continuous surface.}
\label{fig:5}
\end{figure*}

It is worthwhile noticing that Potts or truncated model is unnecessary,
since any disparities change along a ray increases
only once $\tilde{h}_1$ energy per each neighboring sites pair.
It is also important that the inhibit term rejects the solutions with occlusions.
In our model, invisible positions are inherently removed from the beginning.
We do not seek discontinuous surface, but seek continuous surface.
For dashed parts in Figure~\ref{fig:5}, we cannot know
whether some objects surfaces exist or not
from the only one stereo view.

Furthermore, we can think that this problem is not a sites labeling problem,
but a real cross points finding problem from all cross points.
Real cross points make a surface in 3D space.
Graph cuts divide 3D space into two pieces by this surface. 
We will use the term `real cross point' later.

\section{Experiments}
\subsection{Relation between disparity-model and gaze-line model}
We use Tsukuba stereo pair (left: scene1.row3.col1.ppm,
right: scene1.row3.col3.ppm,
$384\times 288$ pixels)
and its ground truth (truedisp.row3.col3.pgm, disparity = $10\dots28$ pixels)
in Middlebury data set.
We introduce the integer coordinates
$(W,H,S)$ for cross points $(g,d,y)$ for simplicity.
$W$ is the horizontal axis which has the left to right direction.
$H$ is the vertical axis which has the up to down direction.
$S$ is the depth axis which has the front to back direction.
$(W,H,S)$ and $(g,d,y)$ are related as
\begin{align*}
W&=g+\textit{offset1}\\
H&=y+\textit{offset2}\\
S&=d+\textit{offset3}.
\end{align*}
Then we denote a pixel in right image as $(x,y)$,
and its disparity as $dis$.
The relation between $(W,H,S)$ and $(x,y,dis)$ is written as follows,
\begin{align*}
W&=x-(lw_{\textit{offset}}+rw_{\textit{offset}})/2+\textit{dis}/2\\
H&=y-h_{\textit{offset}}\\
S&=(lw_{\textit{offset}}-rw_{\textit{offset}})/2-\textit{dis}/2
\end{align*}
and
\begin{align*}
x&=rw_{\textit{offset}}+S+W\\
y&=h_{\textit{offset}}+H\\
\textit{dis}&=lw_{\textit{offset}}-rw_{\textit{offset}}-2 S.
\end{align*}
Where,
$\textit{offset1}$,
$\textit{offset2}$,
$\textit{offset3}$,
$lw_{\textit{offset}}$,
$rw_{\textit{offset}}$,
$h_{\textit{offset}}$
define the center of cuboid area,
but we omit details here.
If $dis$ changes 2, then $S$ changes 1.
Therefore in our model the half of pixels are unused.
This is the weak point of our new site-labeling formulation.
However, since the pixel density of CMOS camera is increasing year by year,
we think this is not significant.
Using this relation, we can translate each other between
ground truth cross points and ground truth disparities.
In dividing-by-2 operations in the above equations, we round away to get integer values.

\subsection{Error count}
For evaluation we define an `error' as follows,
\begin{equation}
\textit{error}=\sum_{s\in \textit{all\_sites}}d(s)=\sum_{d=0}d\times \textit{the\_number\_of\_such\_sites}\label{eq:error}.
\end{equation}
Here, $d(s)$ means difference between the truth depth number
and the estimated depth number at site $s$.

\subsection{Graph cuts using our model}
\begin{figure}[t]
\begin{center}
\fbox{
  \includegraphics[width=0.9\linewidth,clip,trim=8 0 5 3]{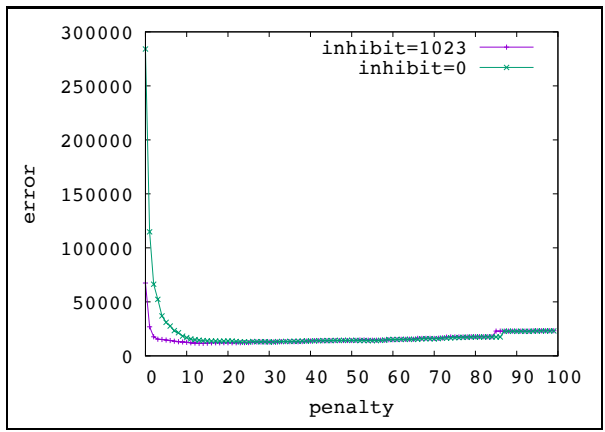}
}
\end{center}
\caption{Relation between the penalty constant and the error (Equation~\ref{eq:error})
at the inhibit constant equals 0 and 1023.}
\label{fig:6}
\end{figure}

\begin{figure}[t]
\begin{center}
\fbox{
  \includegraphics[width=0.8\linewidth,clip,trim=106 97 140 97]{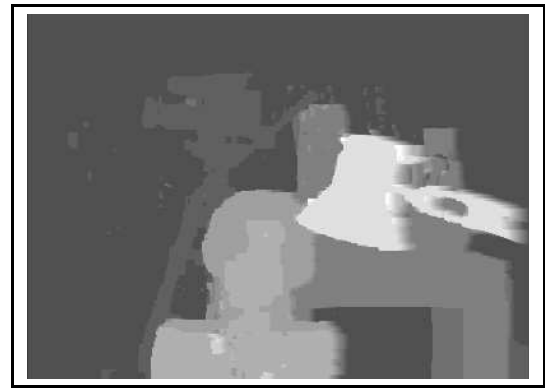}
}
\end{center}
\caption{block\_size=1, exact minimization}
\label{fig:t1}
\end{figure}

We use a cuboid area which includes $372 \times 288 \times 24$ cross points
for Tsukuba stereo pair.
This cuboid cross points are mapped to a graph with $372 \times 288 \times 23 +2=2464130$ nodes
and 19430137 edges. 
We use a simple sum of absolute difference data term as
\[g_v(X_v)=|L_R-R_R|+|L_G-R_G|+|L_B-R_B|,\]
where, $X_v$ means the cross point $(g,d,y)=(x_l,x_r,y)$ in our model,
therefore, $(L_R,L_G,L_B)$ is color of left pixel $(x_l,y)$,
$(R_R,R_G,R_B)$ is color of right pixel $(x_r,y)$.
The color intensity is from 0 to 255.
So, $g_v(X_v)$ varies from 0 to 765.
We use 1023 for the inhibit constant ($\tilde{h}_2$). 1023 is enough big,
since increasing the inhibit constant did not increase max flow in our experiments.
We do not know the theoretical reason to get the best penalty constant value ($\tilde{h}_1$).
We sought it by experiments.
Figure~\ref{fig:6} indicates 14 is experimentally the best.
At this value the error becomes 11578.
Table~\ref{tbl:1} shows details of this case.
The 92\% of sites are equal to the ground truth.
Figure~\ref{fig:t1} is the result disparity image
of our stereo formulation with $\tilde{h}_1=14$,
$\tilde{h}_2=1023$.
Figure~\ref{fig:6} also indicates the difference between with the inhibit constant and
without it.

\begin{table}
\begin{center}
\begin{tabular}{|c|c|c|}
\hline
d		&	number of such sites		&	percentage\\
\hline
\hline
0		&	76502				&	92\%\\
1		&	3766					&	4\%\\
2		&	839					&	1\%\\
3		&	547					&	0\%\\
4		&	98					&	0\%\\
5		&	82					&	0\%\\
6		&	146					&	0\%\\
7		&	46					&	0\%\\
8		&	36					&	0\%\\
9		&	245					&	0\%\\
10$\sim$	&	0					&	0\%\\
\hline
error & 11578 &\\
\hline
\end{tabular}
\end{center}
\caption{Breakdown of the error (Equation~\ref{eq:error}) at inhibit=1023, penalty=14.
d indicates the differences to ground truth.}
\label{tbl:1}
\end{table}

\subsection{GPU graph cuts}
This graph is successfully cut
by BK Max-flow/min-cut code (maxflow-v3.01.zip at vision.csd.uwo.ca/code).
It takes 3823ms at Corei7-5930K 3.5GHz CPU (Table~\ref{tbl:2}).
It is so slow for real time use
that we have developed high speed GPU graph cuts codes.
We have adopted the `push relabel' algorithm
with global label update~\cite{Goldberg1,Goldberg2,Goldberg3}.
The following is our graph\_cut function described in CUDA
(Compute Unified Device Architecture) for NVIDIA GPUs.
Before calling this function,
the `data's from/to the special node S/T are translated to the positive/negative overflows of nodes,
while the `data's between normal nodes are set to the edges.

\small{\begin{verbatim}
void graph_cut(void) {
  int d;
  wave_init<<< grid, block >>>();
  for (int time = 1; time < A; time++)
    wave<<< grid, block >>>(time);
  for ( ; ; ) { // push relabel loop
    bfs_init<<< grid, block >>>();
    for ( ; ; ) {
      bfs_i<<< grid, block >>>();
      d = 0;
      bfs_o<<< grid, block >>>(d);
      if (d == 0) break;
    }
    ovf<<< grid, block >>>(d);
    if (d == 0) break;
    for (int i = 0; i < B; i++)
      push_relabel<<< grid, block >>>();
  }
}
\end{verbatim}}

Where, \verb|grid| $\times$ \verb|block| $=$
372 $\times$ 288 $\times$ 23 $=$ 2464128
and \verb|<<<| \verb|>>>| indicates threads invoking.
Accordingly, 2464128 cuda cores in a GPU are invoked at once.
Before the push relabel loop, we add the `wave front fetch' operation
which we have developed originally.
This increases the graph cut speed twice,
but we omit details here.
It takes 122ms at GTX1080 GPU (Table~\ref{tbl:2}).
It is still slow for real time use.

\subsection{Approximate graph cuts}
\begin{figure}[t]
\begin{center}
\fbox{
  \includegraphics[width=0.8\linewidth,clip,trim=106 97 140 97]{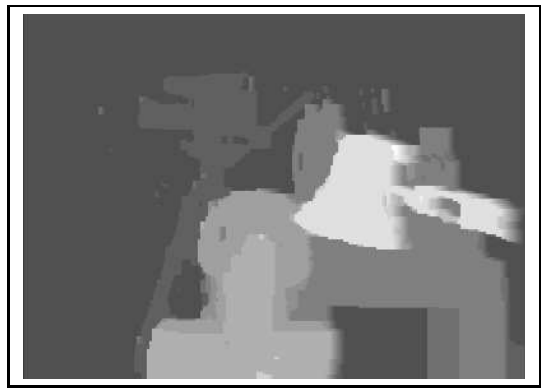}
}
\end{center}
\caption{block\_size=2, level 1 approximation}
\label{fig:t2}
\end{figure}

\begin{figure}[t]
\begin{center}
\fbox{
  \includegraphics[width=0.8\linewidth,clip,trim=106 97 140 97]{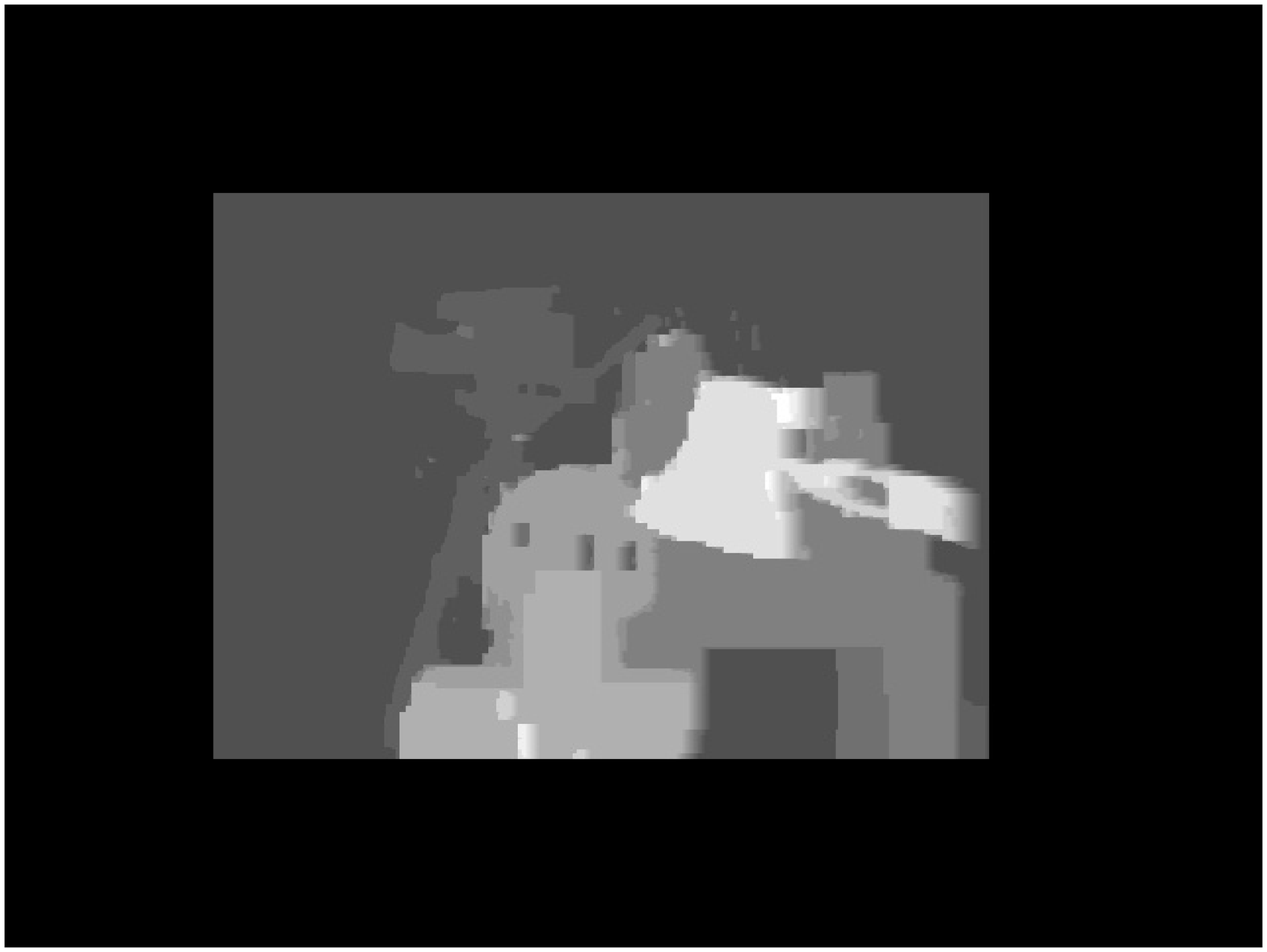}
}
\end{center}
\caption{block\_size=3, level 1 approximation}
\label{fig:t3}
\end{figure}

In order to reduce the processing time,
we have developed the hierarchical graph cuts
which consist of two steps.
In the first step, 2 $\times$ 2 $\times$ 2
or 3 $\times$ 3 $\times$ 3 neighboring
cross points are combined into one delegate point.
The first graph cut determines which delegate points include real cross points.
The second graph cut finds real cross points from the all cross points
which are included in a thin skin 3D area.
We denote this procedure as level 1 approximation.
In the expression 2 $\times$ 2 $\times$ 2
or 3 $\times$ 3 $\times$ 3,
we call 2 or 3 as `block size',
2 $\times$ 2 $\times$ 2
or 3 $\times$ 3 $\times$ 3 as `block'.
These `block' differ from previous \verb|block|.
If the block size equals to 1, the first graph cut find real cross points,
that is, this means exact minimization.
Using level 1 approximation,
the same processing time of 14ms was achieved for both block sizes of 2 and 3
at GTX1080 GPU (Table~\ref{tbl:2}).
Corresponding the result disparity images
are Figure~\ref{fig:t2} and Figure~\ref{fig:t3}, respectively.

\begin{figure}[t]
\begin{center}
\fbox{
  \includegraphics[width=0.8\linewidth,clip,trim=106 97 140 97]{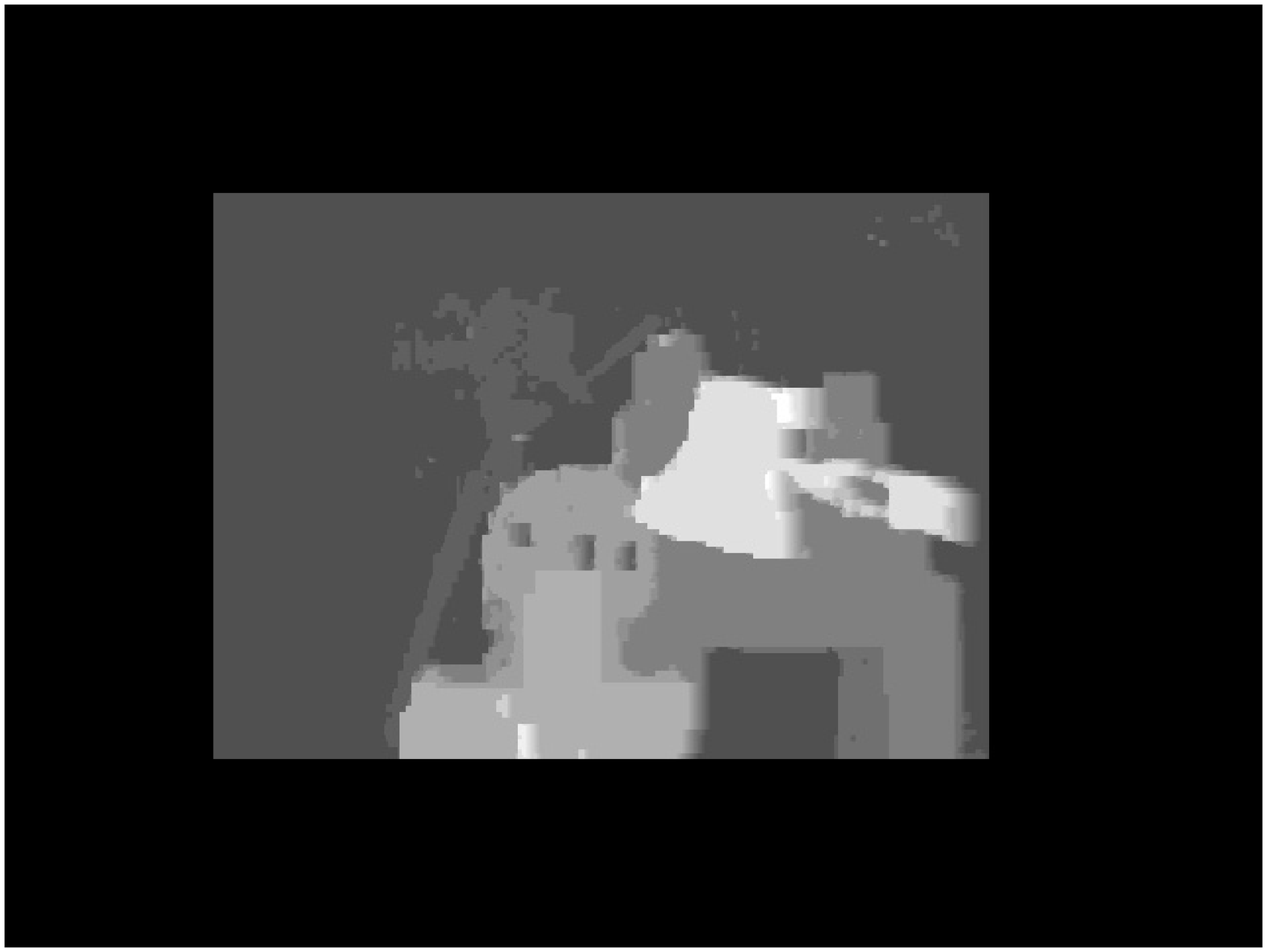}
}
\end{center}
\caption{block\_size=3, level 2 approximation}
\label{fig:t4}
\end{figure}

We have developed further approximation \ie level 2 approximation.
In it, the push relabel loop cuts each block separately,
while the wave front fetch loop deals with the thin skin 3D area as same as in level 1.
This approximation dramatically reduces the processing time.
It takes only 5ms at GTX1080 GPU,
19ms at GTX660 GPU (Table~\ref{tbl:2}).
Corresponding the result disparity image
is Figure~\ref{fig:t4}.
The wave front fetch algorithm is very strong in the early stage
where a lots of overflows exist in the graph,
but very poor for overflows to run out. 
Therefore this combination leads to a good result.

\begin{table}
\begin{center}
\begin{tabular}{|l|c|c|c|c|c|}
\hline
Method       & error & \% &corei7 & gtx1080 & gtx660 \\
\hline
\hline
BK        & 11578 & 92 & 3823ms &   ---   &   ---  \\
l=1 b=1 & 11578 & 92 &   ---  &  122ms  &  472ms \\
l=1 b=2 & 14624 & 91 &   ---  &   14ms  &   83ms \\
l=1 b=3 & 20657 & 88 &   ---  &   14ms  &   86ms \\
l=2 b=3 & 21294 & 88 &   ---  &    5ms  &   19ms \\
\hline
\end{tabular}
\end{center}
\caption{Comparison of the errors (Equation~\ref{eq:error}) and execution times on Tsukuba.
b=1 means exact minimization, l=2 indicates level 2 approximation.
\% indicates the sites percentage of corresponding to ground truth.}
\label{tbl:2}
\end{table}

\section{Conclusion}
The new sites labeling formulation for stereo has been presented. 
We treat left and right image symmetrically
and introduce a novel concept of gaze lines,
which are auxiliary lines defined on
the cross points of rays corresponding to left and right image pixels.
In our formulations, sites are gaze lines and labels are depth numbers.
For this symmetrical formulation,
visibility reasoning is naturally included in the energy function,
and a small smoothness term does not damage the 3D results much.
This makes the minimization step so simple
that we could develop
an approximation method for graph cut itself
which dramatically reduces the processing time.
And now we are noticed that this symmetrical minimization could
perform stereo rectification for the two cameras
without any calibration objects like chess boards~\cite{Zhang}.

\end{document}